\documentclass[journal,twoside,web]{ieeecolor}
\usepackage{tmi}
\usepackage{cite}
\usepackage{amsmath, amssymb}
\usepackage{multirow}
\usepackage{booktabs}
\usepackage{array}
\usepackage{bm}
\usepackage[mathscr]{eucal}
\usepackage{mathtools}
\usepackage{float}
\usepackage{textcomp}

\makeatletter
\def\fnum@figure{\textcolor{subsectioncolor}{\sf Fig.~\thefigure}}
\def\fnum@table{\textcolor{subsectioncolor}{\sf TABLE~\thetable}}
\makeatother

\newtheorem{definition}{Definition}
\newtheorem{theorem}{Theorem}
\newcommand{\floor}[1]{\lfloor #1 \rfloor}
\newcommand{\revision}[1]{\textcolor{black}{#1}}

\newcommand{\scndrev}[1]{\textcolor{black}{#1}}

\def\BibTeX{{\rm B\kern-.05em{\sc i\kern-.025em b}\kern-.08em
    T\kern-.1667em\lower.7ex\hbox{E}\kern-.125emX}}
\begin{document}
\title{Augmented Equivariant Attention Networks for Microscopy Image Transformation}
\author{Yaochen~Xie,~%\IEEEmembership{,}
        Yu~Ding,~%\IEEEmembership{Member,~IEEE,}
        Shuiwang~Ji,~\IEEEmembership{Senior~Member,~IEEE,}

\thanks{Manuscript received xxx; revised xxx. This work is partially supported by AFOSR DDIP program grant FA9550-18-1-0144, NSF grants IIS-1849085, IIS-1908220, and DBI-2028361, and Texas A\&M X-grant program.}
\thanks{Y. Xie is with the Department of Computer Science and Engineering, Texas A\&M University, College Station, TX 77843, USA, e-mail:(ethanycx@tamu.edu).}
\thanks{Y. Ding is with the Wm Michael Barnes'64
Department of Industrial and Systems Engineering, Texas A\&M
University, College Station, TX 77843, USA, e-mail:
(yuding@tamu.edu).}
\thanks{S. Ji is with the Department of Computer
Science and Engineering, Texas A\&M University, College Station, TX
77843, USA, e-mail: (sji@tamu.edu).}}

\maketitle

\begin{abstract}
It is time-consuming and expensive to take high-quality or high-resolution electron microscopy (EM) and fluorescence microscopy (FM) 
images. Taking these images could be even invasive to samples and may damage certain subtleties in the samples after long or intense exposures, often necessary for achieving high-quality or high-resolution in the first place.
Advances in deep learning enable us to perform various types of microscopy image-to-image 
transformation tasks such as image denoising, super-resolution, and segmentation that computationally produce high-quality images from the physically acquired low-quality ones.  When training image-to-image transformation models 
on pairs of experimentally acquired microscopy images, prior models
suffer from performance loss due to their inability to capture inter-image 
dependencies and common features shared among images. Existing methods that take advantage of shared features in image classification tasks cannot be properly applied 
to image transformation tasks because they fail to preserve 
the equivariance property under spatial permutations, something essential 
in image-to-image transformation. To address these
limitations, we propose the augmented equivariant attention networks
(AEANets) with better capability to capture inter-image dependencies, while preserving the equivariance property. The
proposed AEANets captures inter-image dependencies and shared
features via two augmentations on the attention mechanism, which are the shared references and the batch-aware
attention during training. We theoretically derive the equivariance
property of the proposed augmented attention model and
experimentally demonstrate its consistent superiority in both quantitative and visual results over the
baseline methods.
\end{abstract}


\section{Introduction}\label{sec:introduction}
% background: microscopy image transformation, overview of previous methods
\IEEEPARstart{M}icroscopy images of high quality in terms of resolution or noise level are desired to conduct research in various fields such as biomedical science and nanomaterial. However, the capture of high-quality microscopy images is usually at a high cost in budget and time, \revision{and may be infeasible under certain circumstances. This is especially critical for the observation of temporal dynamic or processes of live cells and organics, where exposure time and intensity enable more precise imaging but may reduce the temporal resolution and be harmful to the live cells~\cite{weigert2018content}}. To overcome these drawbacks but yet to produce higher quality images, studies such as microscopy image super-resolution or denoising, aims at computationally producing high-quality microscopy images from the physically acquired low-quality images.

% Deep learning methods... In particular, attention
Deep learning approaches consider the microscopy image super-resolution and denoising as image-to-image transformation tasks, in the sense that pairs of images, of the same size but different noise levels or resolutions, are used to train a deep neural network. One can then apply the trained deep neural network to the low-quality images for predicting their high-quality counterparts. Deep learning approaches have shown success in microscopy image transformation applications on both electron microscopy (EM) and fluorescence microscopy (FM) images, such as content-aware denosing~\cite{weigert2018content}, virtual refocusing~\cite{wu2019three}, and super-resolution~\cite{nehme2018deep, heinrich2017deep, fang2021deep}. In particular, deep learning approaches that involve the self-attention mechanism~\cite{vaswani2017attention} achieve even more promising performance on microscopy image transformation tasks, benefitting from its capability to perform non-local information aggregations and capture long-range dependencies~\cite{wang2020global}.

% Recent studies show the importance of shared features
Recently, several studies~\cite{qian2020effective, liu2020global, yuan2019learning} have shown or indicated the importance of capturing the inter-images dependencies and shared features among images, due to the uniqueness of microscopy images. For example, \cite{qian2020effective} contemplate the question of what  training strategy is the best for their EM super-resolution model and compare two strategies.  The pooled-training uses all training image-pairs to train a single model and perform prediction on all testing areas, whereas the self-training trains a dedicated model for each unique testing area with the training image-pair from the same samples of the corresponding training area. For each model trained, the self-training uses far fewer image pairs than those used in pooled-training, yet \cite{qian2020effective} showed that the models trained with self-training generally yield higher performance than pooled-training, which on the surface appears to contradict the conventional wisdom in deep learning that a model trained with more data should generally perform no worse than that with fewer data. We attribute the reduced performance of pooled-training to the models' inability to fully utilize the additional information provided in pooled-training. More importantly, we observe that this lack of capability is not unique to the deep learning methods tested by \cite{qian2020effective}, but rather common among most existing deep learning methods. The methods are inadequate in terms of capturing inter-image dependencies and shared features among training image-pairs, which are prerequisite for ensuring the pooled-training strategy to do better.

% recently, attention mechanism, also known as the transformer
To address the above issue, an existing attention-based approach has been proposed to endow neural networks with the capability to capture shared features among training instances. That is, to include the attention mechanism with learnable query as an augmentation to the original self-attention mechanism. Such an approach has been explored and commonly used in natural language processing (NLP)~\cite{yang2016hierarchical} and graph neural networks (GNNs)~\cite{li2015gated}. In the image domain, \cite{liu2020global, yuan2019learning} have also attempted to include the attention mechanism with learnable query to capture the inter-image dependencies and the common features shared among images. 

However, we argue that the attention mechanism with learnable query can be inappropriate in the image-to-image transformation tasks, leading to potential performance reduction of the model due to the lacking of an essential property, the spatially permutation equivariant. When we perform spatial permutation such as rotation to an input image, the output image is desired to be permuted accordingly. For typical convolution-based deep models, such an equivariant property can be naturally learned or enforced by performing data augmentation such as rotation and flipping. However, involving the attention mechanism with learnable query makes such a property unsatisfied in an image-to-image transformation model and unable to be learned, unless constant values are output by the attention operator at all spatial locations. Consequently, although shared features can be captured, image-to-image transformation models involving the attention mechanism with learnable query suffers from performance loss due to the lack of permutation equivariance.

% Our contribution
Motivated by both the desire to utilize inter-image dependencies and overcome the limitations of attention mechanisms with learnable query, we propose the augmented attention models with two components, the attention mechanism with shared references and the batch-aware attention applied in training. The resulting new attention model is referred to as the Augmented Equivariant Attention Networks (AEANets), whose attention block preserves the equivariance to any spatial permutations and can capture the inter-image dependencies and common features among images. We conduct experiments to evaluate the performance and effectiveness of the proposed AEANets. Quantitative results show that our AEANets significantly outperform the baselines on three microscopy image transformation tasks, i.e., super-resolution, denoising, projection, and segmentation, for both various types of biomedical images. We also demonstrate visually that AEANets produce better 3D-to-2D projection and super-resolution images compared to the respective baseline methods.

%\enlargethispage{12pt}

\section{Preliminaries and Related Studies}
In this section, we introduce the self-attention mechanism and related studies that apply the self-attention mechanism or its variations with learned query.

\subsection{The Self-Attention Mechanism}\label{sec:selfatt}
The self-attention mechanism \cite{vaswani2017attention} has been widely applied to deep learning models in natural language processing (NLP) \cite{devlin2018bert} and computer vision \cite{wang2018non}. Compared to local operations such as convolutions that can only aggregate information locally, the self-attention mechanism is able to incorporate global information. Given an input feature map, the self-attention mechanism computes the relevance between every two locations on the feature map and aggregations information from one location to another according to the relevance. The self-attention mechanism hence endows neural networks the capability to capture long-range dependencies.

% can be applied to higher dimensions, but we discuss the 1D case.
The self-attention mechanism can be applied to feature maps $\boldsymbol{\mathscr{X}}\in\mathbb{R}^{s_1\times\cdots\times s_k\times c}$ with any $k\ge1$ where $k$ denotes the number of spatial dimensions, $s_i$ denotes the spatial size along the $i$-th dimension and $c$ denotes the number of features. For example, in a 2D image case, the self-attention mechanism is applied on the input $\boldsymbol{\mathscr{X}}\in\mathbb{R}^{w\times h\times c}$ where $w$ and $h$ denote the width and height of the image. Without loss of generality, we describe how the self-attention operator is performed in the 1D case ($k=1$), where there is only one spatial dimension. For higher-dimensional cases, the spatial dimensions can be unfolded into one dimension $s=s_1s_2\cdots s_k$ before being given to the self-attention operator. The output of the self-attention operator can then be folded back to the original shape as the final output.

In the 1D case, the self-attention operator takes as input a matrix $\bm X\in\mathbb{R}^{s\times c}$ representing the features of a sequence, where $s$ denotes its spatial dimension (\emph{i.e.}, the length of the sequence or the spatial dimensions of unfolded images) and $c$ denotes its feature dimension. The self-attention operator firstly computes three matrices, \emph{i.e.}, the query $\bm Q$, the key $\bm K$ and the value $\bm V$, by performing convolutions with kernel size of 1, to the input matrix $\bm X$. Formally, $\bm Q = q(\bm X)\in\mathbb{R}^{s\times c_1}$, $\bm K = k(\bm X)\in\mathbb{R}^{s\times c_1}$ and $\bm V = v(\bm X)\in\mathbb{R}^{s\times c_2}$, where $q(\cdot),k(\cdot),v(\cdot)$ are three independent projections. Then, the output $\bm Y$ of the attention operator is computed by
\begin{equation}\label{eq:selfatt}
\bm Y = \mbox{Normalize}(\bm Q\cdot\bm K^T)\cdot\bm V\in\mathbb{R}^{s\times c_2}.
\end{equation}
The function $\mbox{Normalize}(\cdot)$ performs a normalization on the attention map $\bm Q\cdot\bm K^T$ so that the values on the output $\bm Y$ will not scale with the spatial size. Commonly used $\mbox{Normalize}(\cdot)$ functions includes $\mbox{Softmax}(\cdot)$ and the division by the spatial size of $\bm K$, \emph{i.e.},
\begin{equation}\label{eq:norm}
\mbox{Normalize}(\bm Q\cdot\bm K^T)=\frac{1}{s}(\bm Q\cdot\bm K^T).
\end{equation}
In this work, we use the normalization function in Equation~(\ref{eq:norm}). For clear comparisons, we use this type of normalization when describing all variations of the self-attention mechanisms in the rest of the paper.

Note that although the spatial sizes of $\bm Q,\bm K,\bm V$ are the same in the self-attention mechanism, the spatial size $s$ of the output is determined by the spatial size of query $\bm Q$. In addition, the feature size $c_2$ of the output is determined by the feature size of value $\bm V$.

\subsection{Attention Mechanism with Learned Query}\label{sec:learnq}
% introduction
A common variation of the attention mechanism is to directly learn the values in the query matrix $\bm Q$. In this case, the query $\bm Q$ does not depend on the input $\bm X$.  The attention mechanism with a learnable query is commonly used in NLP \cite{yang2016hierarchical} and graph neural networks (GNNs) \cite{li2015gated}.
In certain domains such as biomedical image and nanoparticles, different images from one dataset usually share similar patterns and common features, such as microscopy images captured from different parts of tissue or tissues of the same type. The power of the learnable query has also been explored by previous studies in the biomedical image domain~\cite{yuan2019learning,liu2020global}. In these cases, such a variation of the attention mechanism allows the networks to capture common features from all input images during training since the query is independent of the input and is shared by all input images.

% method description
Instead of computing the query from $\bm X$, the attention mechanism can learn a matrix $\bm Q\in\mathbb{R}^{s_q\times c_1}$ independently of $\bm X$. The other computations in the attention mechanism with learnable query is then the same as the self-attention mechanism. Formally, with $\bm K=K(\bm X)\in\mathbb{R}^{s\times c_1}$, $\bm V=V(\bm X)\in\mathbb{R}^{s\times c_2}$ and a directly learned matrix $\bm Q\in\mathbb{R}^{s_q\times c_1}$,
\begin{equation}
    \bm Y = \mbox{Normalize}(\bm Q\cdot\bm K^T)\cdot\bm V\in\mathbb{R}^{s_q\times c_2}.
\end{equation}

Since the spatial size of the output $\bm Y$ is determined by the spatial size of $\bm Q$, the output size of the attention mechanism with learned query is no longer related to the spatial size of $\bm X$. As a result, when the attention mechanism with learned query is included in the neural network, the size of the output of the network is usually fixed.

\section{Model Augmentation with Shared References}
Previous studies~\cite{yuan2019learning, liu2020global} have explored different approaches to include a learnable query in the attention operator to capture common features among different images. Such attention operators with learnable query have been shown to bring a promising performance boost, especially in NLP and image classification tasks. However, the attention operators with learnable query can, on the contrary, limit the performance of models for image-to-image transformation tasks such as image super-resolution, as such operators are not able to preserve an essential property required by the image-to-image transformation models, \emph{i.e.}, the equivariance to spatial permutations. \scndrev{Note that such properties are not changed by increasing network depth, modifying architectures other then the attention operator, or simply applying data augmentations. Therefore, the issue cannot be addressed by these approaches.}

In this section, we analyze the equivariance property in subsection~\ref{sec:property}, and show that such property is violated when the attention mechanism includes learned query in subsection~\ref{sec:property}. Based on our analysis, we propose in subsection~\ref{sec:sr} the attention operator with shared references that are able to capture common features among images without violating the equivariance property.

\subsection{Equivariance and Invariance to Spatial Permutations}\label{sec:property}

The spatial permutation includes a group of transformations to be applied to images. It is performed by permuting the spatial locations of any number of pixels or voxels in an image. Some common examples of spatial permutation include the rotation, the flipping and the shifting of an image. The equivariance to the spatial permutation is a property of an operator or a model such that applying a spatial permutation to the input of the operator or the model results in an equivalent effect of applying the same spatial permutation to the output. 
On the contrary, if an operator is invariant to spatial permutations, then its output remains unchanged when permuting the input.
We provide formal definitions of the spatial permutation, the equivariance and invariance property below.
\begin{definition}\label{def:sp}
Consider an image or feature map $\bm X\in \mathbb{R}^{s\times c}$, where $s$ denotes the spatial dimension and $c$ denotes the number of features. Let $\pi$ denote a permutation of $s$ elements. We call a transformation $\mathcal{T}_\pi: \mathbb{R}^{s\times c}\to \mathbb{R}^{s\times c}$ a spatial permutation if $\mathcal{T}_\pi(\bm X)=P_\pi\bm X$, where $P_\pi\in\mathbb{R}^{s\times s}$ denotes the permutation matrix associated with $\pi$, defined as
%\begin{equation}
    $P_\pi =
    \left[
        \bm{e}_{\pi(1)},
        \bm{e}_{\pi(2)},
        \cdots,
        \bm{e}_{\pi(s)}
    \right]^T,$
%\end{equation}
and $\bm{e}_i$ is a one-hot vector of length $s$ with its $i$-th element being 1.
\end{definition}

\begin{definition}\label{def:inveq}
We call an operator $A: \mathbb{R}^{s\times c_1}\to \mathbb{R}^{s\times c_2}$ to be spatially permutation equivariant if $\mathcal{T}_\pi(A(\bm X)) = A(\mathcal{T}_\pi(\bm X))$ for any $X$ and any spatial permutation $\mathcal{T}_\pi$. In addition, an operator $A: \mathbb{R}^{s\times c_1}\to \mathbb{R}^{s\times c_2}$ is spatially permutation invariant if $A(\mathcal{T}_\pi(\bm X)) = A(\bm X)$ for any $X$ and any spatial permutation $\mathcal{T}_\pi$.
\end{definition}

\begin{figure}[h]
    \centering
    \includegraphics[width=0.48\textwidth]{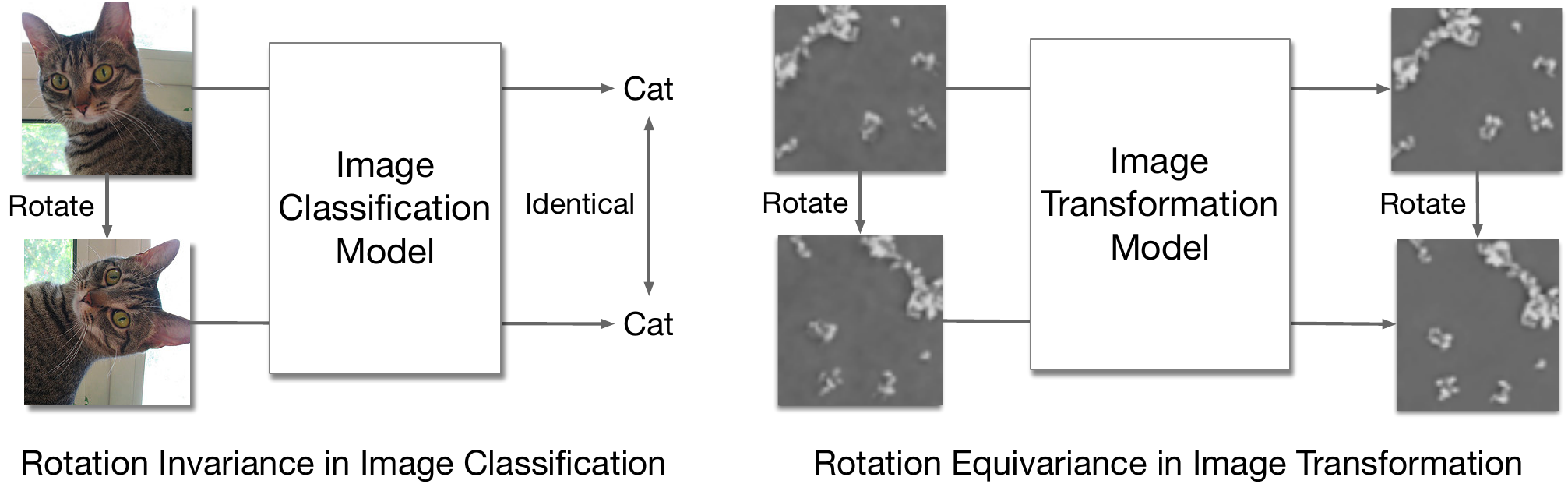}
    \caption{Examples that compare the invariance property with the equivariance property. Tasks such as classification require spatial permutation (e.g., rotation) invariant models, where applying the permutation to the input does not change the output of the model. On the contrary, the image transformation tasks require spatial permutation (e.g., rotation) equivariant models, where applying the permutation to the input leads to the same permutation applied to the output of the model.}
    \label{fig:inv_eqv}
\end{figure}

% We now compare the two properties. 
We argue that while the image classification models could benefit from the invariance to spatial permutations according to previous studies~\cite{marcos2016learning, zhang2019making}, and an image-to-image transformation model requires the equivariance to spatial permutations, as shown in Figure~\ref{fig:inv_eqv}. Detailed discussions about the properties are provided in Appendix~\ref{supp:disc}.

\subsection{Spatial Permutation Properties of Attention Operators}

We now analyze the properties of the two types of attention operators regarding the spatial permutation using the same notations as in Section~\ref{sec:selfatt} and Section~\ref{sec:learnq}. Intuitively, when a spatial permutation is performed on the input of an attention operator, the corresponding permutation made on the key $\bm K$ and the value $\bm V$ does not result in any difference in the output, as long as the same permutation is applied to both $\bm K$ and $\bm V$. In fact, the order of spatial locations on the output feature map is determined by the spatial locations on the query $\bm Q$. Hence the attention operator is permutation equivariant as long as $\bm Q$ is obtained from $\bm X$.

For simplicity, we denote $A_s$ a self-attention operator and $A_{\bm Q}$ an attention operator with learned query. The outputs $\bm Y$ of the two operators are therefore equal to $A_s(\bm X)$ and $A_{\bm Q}(\bm X)$. We show that the following theorem holds.

\begin{theorem}\label{th:1}
A self-attention operator $A_s$ is permutation equivariant while an attention operator with learned query $A_{\bm Q}$ is permutation invariant. In particular, letting $\bm X$ denote the input matrix and $\mathcal{T}$ denotes any spatial permutation, we have
$$A_s(\mathcal{T}_\pi(\bm X))=\mathcal{T}_\pi(A_s(\bm X)),$$
and
$$A_{\bm Q}(\mathcal{T}_\pi(\bm X))=A_{\bm Q}(\bm X).$$
\end{theorem}

The proof of Theorem~\ref{th:1} is provided in Appendix~\ref{supp:proof1}. Note that we prove the above theorem with the normalization of division by spatial size of $\bm K$. The theorem still holds when the $\mbox{Softmax}$ is applied for normalization, the proof of which can be found in~\cite{Ji:attn}.

% Note that the above theorem still holds when the $\mbox{Softmax}$ is applied for normalization, the proof of which can be found in~\cite{Ji:attn}.

\subsection{Augmented Attention with Shared References}\label{sec:sr}
\begin{figure*}[th]
\centering
\includegraphics[width=\textwidth]{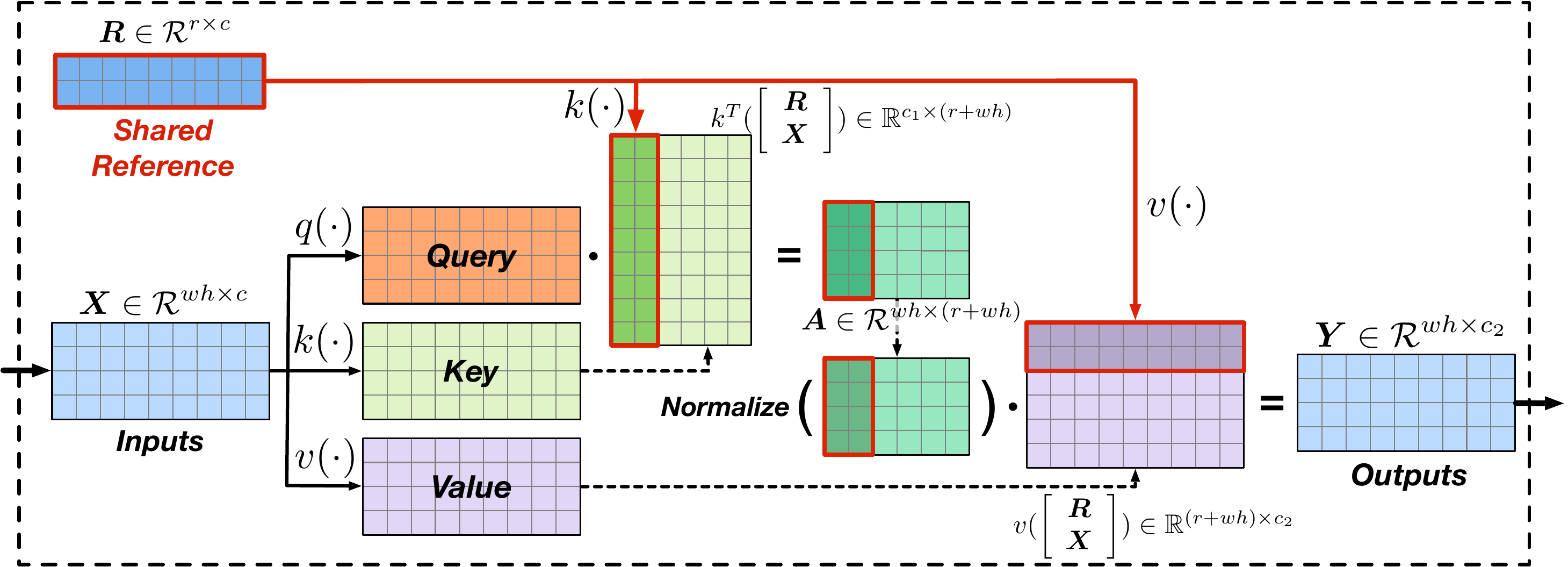}
\caption{Comparison between the original self-attention (top) and proposed attention operator with shared references (bottom). Given the flattened input matrix $\bm X$ of shape [width $\times$ height, channels], the proposed attention operator train a reference matrix of shape [reference size, channels], which is used to compute the key and value. The highlighted parts in the key and value are related to the shared references and the rest parts are related to the input image. The spatial size of the output is the same as the spatial size of the query.}\label{fig:sr}
\end{figure*}

We have shown that an image-to-image transformation model requires equivariance to spatial permutations while the attention operator with learned query is permutation invariant. Although the attention operator with learned query endows models the capability to capture common features among images, the invariance property makes it inappropriate to be applied in image-to-image transformation tasks. In order to endow the attention operator the capability to capture common features among images without losing the equivariance property, we propose an attention operator augmented with learnable shared references, as opposed to shared query. The shared references are represented by a matrix consisting of learnable variables and are augmented to the key and value matrices along the spatial dimensions.

To be concrete, given the flattened input feature map $\bm X\in \mathbb{R}^{wh\times c}$ where the width $w$ and height $h$ are flattened into one dimension and $c$ is the number of features, the shared references are represented by a learnable matrix $\bm R\in \mathbb{R}^{r\times c}$, where $r$ is the size of the shared references as a hyper-parameter. The learned shared references is projected by $k(\cdot)$ and $v(\cdot)$ into the same space of key and value. The computation of an attention operator augmented with shared references $A_{\bm R}$ can be formally expressed as
\begin{equation}
\begin{split}
    \tilde{\bm X} &= \left[
\begin{array}{c}
\bm R\\
\bm X
\end{array} \right]\in \mathbb{R}^{(r+wh)\times c},\\
    A_{\bm R}(\bm X) &= \frac{1}{r+wh} (q(\bm X)\cdot k^T(\tilde{\bm X}))\cdot v(\tilde{\bm X}),
\end{split}
\end{equation}
where $k^T(\tilde{\bm X})$ denotes the transposed key $k(\tilde{\bm X})$. Note that the key $\tilde{\bm K}=k(\tilde{\bm X})$ and value $\tilde{\bm V}=v(\tilde{\bm X})$ are computed from $\tilde{\bm X}$, while the query $\bm Q$ is computed from $\bm X$. The operator with shared references is illustrated in Figure~\ref{fig:sr}. We now show the property of the proposed attention operator $A_{\bm R}$ in the following theorem.
\begin{theorem}\label{th:2}
The proposed augmented attention operator with shared references $A_{\bm R}$ is spatially permutation equivariant, \emph{i.e.},
$$A_{\bm R}(\mathcal{T}_\pi(\bm X))=\mathcal{T}_\pi(A_{\bm R}(\bm X)).$$
\end{theorem}

The proof of Theorem~\ref{th:2} is provided in Appendix~\ref{supp:proof2}.

Compared with the original self-attention operators in which the key and value matrices are fully based on the input, the key and value of $A_{\bm R}$ contain additional information about the features shared by all images in the dataset. The learning process of the shared reference is to distill common features from images in the entire training data. Each spatial location on an input instance aggregates information not only globally from the input instance itself, but also from the distilled references shared by all the input images.

\section{Model Augmentation with Batch-Aware Training}

A common strategy to train a deep model is to feed a mini-batch of images to the network at each training step. The mini-batch is referred to as batch in the following paragraphs for short. When a self-attention operator is included, the operator processes the batch at an instance level. Given an input batch of feature maps $\{\bm X_1, \cdots, \bm X_N\}$, where $N$ is the batch size, the self-attention operator computes the outputs individually for each instance in the batch, \emph{i.e.}, $\bm Y_i = A_s(\bm X_i)$ for $i=1,\cdots,N$.

In the case where images in a dataset share similar patterns, the performance of a deep model can further benefit by incorporating cross-images dependencies. In this case, the learning of such dependencies across images in a batch can be of great importance. Due to the non-local property, the attention operators can be extended from the instance level to a batch level in order to learn the correlations across images in a batch. Formally, we define an augmented batch-aware operator $A_{batch}$ such that
$$\bm Y_i = A_{{batch}}(\bm X_i; {\bm X_1, \cdots, \bm X_N}), i=1,\cdots,N.$$
In this case, the computation of each output instance is aware of the other instances in the current batch. In order to realize such an augmentation, we propose a training strategy with this batch-aware attention, where the key and value cover all the images in the training batch. That is,
\begin{equation}
    \begin{split}
        &A_{batch}(\bm X_i; \bm X_1, \cdots, \bm X_N) = \\ &\frac{1}{Nwh}q(\bm X_i)\cdot k^T\left(
        \left[\begin{array}{c}\bm X_1\\ \vdots \\\bm X_N\end{array} \right]
        \right)\cdot v\left(\left[\begin{array}{c}\bm X_1\\ \vdots \\\bm X_N\end{array} \right]\right), \\
    \end{split}
\end{equation}
where $k(\cdot)$ and $v(\cdot)$ are the projections defined in the original self-attention.

The proposed batch-aware attention aggregates information from the entire batch based on the correlation across images for each input location. Since each batch is uniformly sampled from the entire dataset, the aggregation from batches can estimate the information aggregation from the entire dataset. The weights of the projections $q(\cdot),k(\cdot),v(\cdot)$ in batch-aware attention and the attention with shared references are shared during training. The purpose of including the batch-aware attention in training is to help the distill of shared references and learn better projections in attention by using cross-image dependencies. In our experiments, we empirically show that additional performance improvements can be achieved by including the augmented batch-aware attention in training.

\section{Augmented Equivariant Attention Networks}
\subsection{The Augmented Equivariant Attention Block}
The proposed augmented equivariant attention block consists of a branch for the attention operator $A_{\bm R}$ with shared references and a branch for the batch-aware attention $A_{batch}$. In the attention block, the weights in each projection of $q(\cdot), k(\cdot)$ and $v(\cdot)$ are shared by the two attention operators $A_{\bm R}$ and $A_{batch}$. As the attention block performs differently during training and prediction, we individually describe how it works during training and prediction.

\begin{figure}[t]
    \centering
    \includegraphics[width=0.48\textwidth]{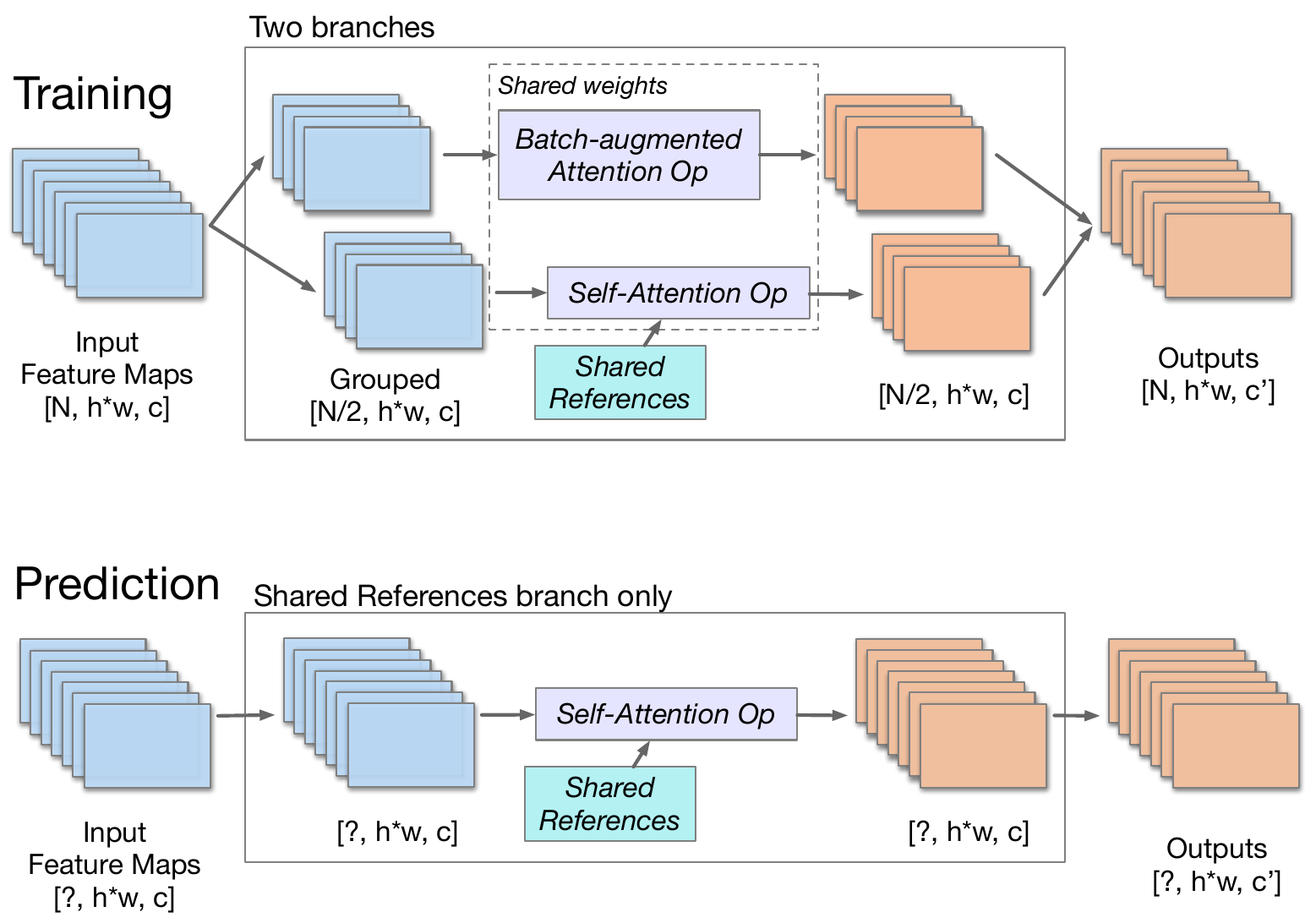}
    \caption{The proposed attention block. During training (top), both branches are used. The input batch are splitted into two groups and passed to the two operators. The outputs of the two operators are then merged together. For prediction (bottom), only the branch with shared references is used.}
    \label{fig:block}
    \vspace{-8pt}
\end{figure}

During training, both the $A_{\bm R}$ and $A_{batch}$ are used. Given an input batch $\{\bm X_1, \cdots, \bm X_N\}$ to the block, the batch is evenly split into two groups $\{\bm X_1, \cdots, \bm X_{\floor{N/2}}\}$ and $\{\bm X_{\floor{N/2}+1}, \cdots, \bm X_N\}$ as the inputs to the two branches. The outputs of the two branches are then merged back into a complete batch. While the batched data in the two branches are separate, the parameters of the two branches are shared.

Once the network is trained, the parameters in the two operators are fixed and are used by the attention operation with shared references during prediction. The batch-aware attention is excluded during prediction since there is not necessarily a batch input during prediction. In other words, the branch that contains the batch-aware attention operator is disabled, and all input data flow into the $A_{\bm R}$ branch. In spite of the exclusion of the batch-aware attention, the existence of shared references distilled from training images allows the model to still utilize the dependencies between training images and the given input image for prediction. Figure~\ref{fig:block} illustrates how the block works during training and prediction.

\subsection{Network Architecture}
Recent studies have shown that the U-Net~\cite{ronneberger2015u} architecture achieves promising performance in many image transformation tasks, especially for microscopy images~\cite{liu2020global,zwangUnetAAAI20,Zeng:Bioinfo17}. In this work, we use the U-Net as the base network architecture of our model. To be specific, we use a U-Net with a depth of 3 (including two down-sampling operations and up-sampling operations, respectively). The skip-connections in the U-Net are merged into the up-sampling path by concatenation.

To enable the use of the proposed augmented attention within the U-Net architecture, we follow~\cite{wang2020global} to (1) include a residual connection in the attention block by addition and (2) substitute the bottom block in the original U-Net with our proposed attention block. In addition, the proposed attention blocks can be also applied as up-sampling blocks by performing up-sampling to the query $\bm Q$~\cite{wang2020global}. The overall architecture of the proposed network is shown in Appendix~\ref{supp:network}.

Note that although we use the augmented attention blocks in the U-Net architecture in this study, it can be inserted into any other deep architectures, thereby capturing non-local and cross-image dependencies and the common features shared by the entire dataset.

\section{Experimental Results}
We evaluate the proposed AEANet for different microscopy image transformation tasks on three microscopy image datasets captured by different instruments. The datasets are the Paired Electron Microscopy (EM) Image Dataset~\cite{qian2020effective}, the Planaria dataset for 3D image denoising; and the Flywing dataset for 3D image projection---the latter two datasets are from CARE~\cite{weigert2018content} and captured by fluorescence microscopy. For all the three datasets, the low-quality images and their high-quality counterparts are physically captured. For each of the three datasets, we follow baseline methods for their experimental settings including training-test split and basic neural network configurations for fair comparisons. For all microscopy image transformation tasks, we use two evaluation metrics, the structural similarity index measure (SSIM) and peak signal-to-noise ratio (PSNR), calculated between the prediction and the high-quality images (ground truth). \revision{We additionally conduct experiments on the 3D segmentation task with brain Magnetic Resonance Images (MRI) to demonstrate broader application scenarios for the proposed methods. We summarize the implementation details and configurations of our methods for individual experiments in Appendix~\ref{supp:config}.}

\subsection{The Paired EM Image Dataset}
We first train and evaluate our model on the publicly available\footnote{The paired EM images dataset is available for public access at {{https://aml.engr.tamu.edu/2001/09/01/publications/}} (then go to J74).} paired EM images dataset~\cite{qian2020effective}. The paired image dataset consists of 22 pairs of LR and HR nanoimages of size $1,280\times944$. The LR and HR nanoimages are captured by the same scanning electron microscope (SEM) at two different magnifications. Specially, the HR image is two times zoomed-in from the LR image and the field of view (FOV) of the HR image is covered by the FOV of the LR image, \emph{i.e.}, the HR image corresponds to a $1/4$ sub-area of the LR image. Several preprocessing steps are performed on the original LR and HR images to build our dataset for training and testing. We first perform the Random Sample Consensus~\cite{fischler1981random} (RANSAC) algorithm to register the HR images in the corresponding areas in the LR images, based on the ORB features~\cite{rublee2011orb}. We select the registered area in each LR image and use a bicubic interpolation to upsample the selected LR subareas to be of the same size of the HR images, \emph{i.e.}, $1,280\times 944$. Through the preprocessing, the resulting LR and HR images refer to the same area but are in different resolutions.

\subsubsection{Training-Test Split and Training Strategies}
\revision{As described in Section~\ref{sec:introduction}, \cite{qian2020effective} studied two training strategies, self-training and pooled-training, for the EM image super-resolution. While both self-training and pooled-training are practical in real scenarios of super-resolution, we only focus on the pooled-training, which is more common in machine learning studies, in our evaluation. In particular, the pooled-training trains a single model on all the 22 image pairs. With the splitting of the original images into $3\times 4$ smaller sub-images, in the pooled-training, the 22 image pairs become a total of 198 sub-image pairs for training and 66 sub-image pairs for test. The evaluation metrics are computed on each testing sub-image and then averaged.}

\subsubsection{Evaluation Results}
We evaluate our method and compare it with an array of deep learning-based baselines.  In addition to three deep learning methods compared in \cite{qian2020effective}, \emph{i.e.}, VDSR~\cite{kim2016accurate}, RCAN~\cite{zhang2018image}, and EDSR~\cite{lim2017enhanced}, we further include the original U-Net~\cite{ronneberger2015u} and GVTNets~\cite{wang2020global}, the current state-of-the-art model for microscopy image transformation, in this comparison study. The U-Net, GVTNets, and our proposed AEANets use the same network architecture setting except for the attention blocks. We also include the SOTA non-deep-learning-based method, which is the paired LB-NLM~\cite{qian2020effective} in our baselines. We show in Table~\ref{tab:main_results} the averaged improvements in terms of the two metrics, as compared to the input LR image (i.e., after bicubic interpolation). The improvements in the two metrics are denoted as $\Delta$PSNR and $\Delta$SSIM.

The results show that the AEANets model with pooled-training significantly outperforms the baselines with the same training strategy. More importantly, the performance of AEANets with pooled-training is better than the baseline models with self-training, indicating that the self-training is no longer required for our proposed AEANets. In other words, AEANets can be used more efficiently, leading to better performance, and can be applied in broader scenarios where self-training may not be applicable.

While the improvement in terms of PNSR is moderate, the improvement in terms of SSIM is much more remarkable, a 70\% increase as compared to the best of the three deep learning methods originally used in \cite{qian2020effective}. Recall that SSIM measures how far away an image is from the HR image and a higher SSIM suggests a better capability of the resulting image to show finer details. The improvement in SSIM bears important practical implication for material characterization.

To see the implication from a different angle, consider the image quality improvement only in the foreground of the images.
Not surprisingly, material scientists are more interested in the nanomaterial clusters (foreground) than the host material (background) in their applications.  To separate the foreground and background, we follow~\cite{qian2020effective} and perform Otsu’s algorithm on each testing patch. We compute the improvements in PSNR on the foreground and background for the following methods: VDSR, SRSW, Paired LB-NLM and AEANets.  The outcome is shown in Table~\ref{tab:main_results} (right). The results demonstrate that although VDSR with self-training achieves a higher PSNR in the background, AEANets outperform the three aforementioned methods for the foreground.  This outcome reinforces the advantage of AEANets shown in the SSIM comparison.

In terms of the foreground-background difference, \cite{qian2020effective} in fact commented that ``\emph{It is
apparent that all these methods [those included in their paper] denoise the background much
more than they enhance the foreground}.'' While AEANets still see a greater PSNR improvement over its background, its foreground-background performance gap is the smallest among the four alternatives in Table~\ref{tab:main_results}  (right). The practical implication is that AEANets are a better tool for material characterization.

\begin{table*}[th]
\centering
\caption{\textit{Left}: Comparison of performance quantified by image quality improvement in terms of PSNR and SSIM, among our methods and the baseline models. Bold numbers are the highest compared among results in both training strategies. \textit{Right}: Improvements in PSNR on foreground (nanomaterial clusters) and background (host material) individually. ``Self'' in the brackets after method names refers to self-training.}\label{tab:main_results}
\begin{tabular}{lcccc}
\toprule
& \multicolumn{2}{c}{Pooled-training} & \multicolumn{2}{c}{Self-training} \\
Methods                     & $\Delta$PSNR (dB) & $\Delta$SSIM & $\Delta$PSNR (dB) & $\Delta$SSIM \\ \hline

VDSR~\cite{kim2016accurate}& 1.25         & 0.047        & 2.07         & 0.051        \\
RCAN~\cite{zhang2018image} & 1.59         & 0.051        & 2.07         & 0.050        \\
EDSR~\cite{lim2017enhanced}& 1.35         & 0.051        & 2.06         & 0.052        \\
Paired LB-NLM~\cite{qian2020effective} & 0.78         & 0.031        & 1.67         & 0.037        \\ \cline{1-5}

U-Net~\cite{ronneberger2015u}          & 1.46         & 0.074        & --           & --            \\
GVTNet~\cite{wang2020global}           & 1.87         & 0.086        & --           & --            \\ 
\revision{Learned Query}                          & 1.64         & 0.084        & --           & --            \\ \cline{1-5}

AEANet (Ours)                 & \textbf{2.10}         & \textbf{0.087}        & --         & --        \\ \bottomrule
\end{tabular}
\quad\quad
\begin{tabular}{lcc}
\toprule
Methods                     & Foreground & Background \\ \hline
VDSR (Self)           & 0.97         & \textbf{2.83}\\
SRSW (Self)           & -0.25        & 2.15         \\
Paired LB-NLM (Self)  & 0.23         & 2.65         \\ \cline{1-3}
Learned Query (Pooled)               & 0.75 & 1.95        \\
AEANet (Pooled)               & \textbf{1.15}& 2.42        \\ \bottomrule
\end{tabular}
\end{table*}

\begin{figure}[t]
    \centering
    \includegraphics[width=0.48\textwidth]{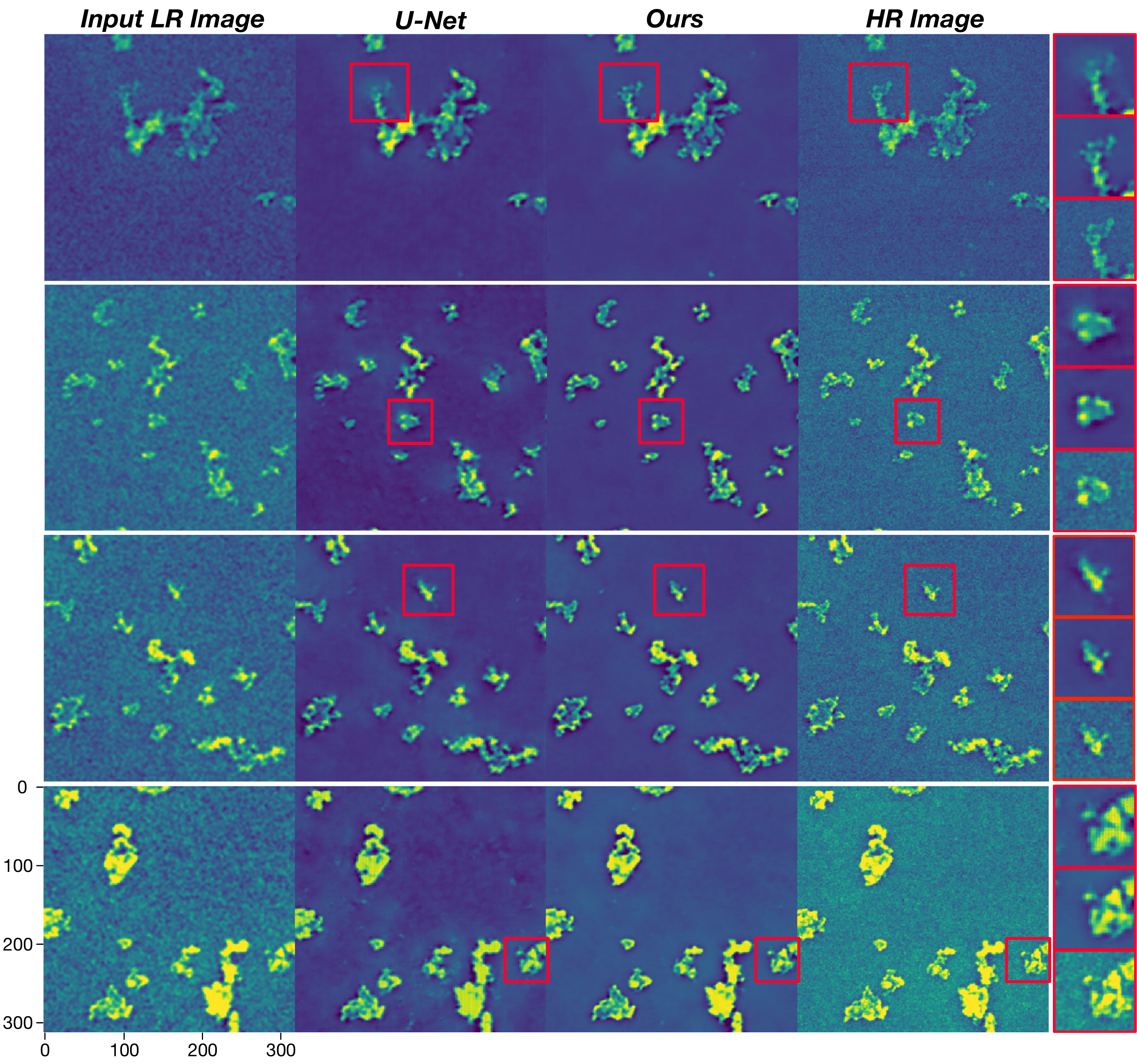}
    \caption{Visualization of the output of super-resolution models on four testing patches. From left to right, the columns are input LR images, outputs of the U-Net, outputs of our model and the ground truth HR images). We select some areas to zoom in for a better view. From top to bottom, smaller patches on the right are zoomed-in from U-Net, our model and the HR images.}
    \label{fig:vis_main}
    \vspace{-8pt}
\end{figure}

We also include some visualization results to compare the super-resolution performance of AEANets with that of U-Net. The visualization shows that the predictions from AEANets have clearer edges of the nanomaterial clusters. It is also worth mentioning that the deep learning-based methods, including U-Net and AEANets, have an additional denoising effect in the background. We believe that the effect is due to the noise in the LR image and HR image belongs to the same noise distribution but are independently sampled, which satisfies the requirements of the Noise2Noise~\cite{lehtinen2018noise2noise} denoising. The denoising effect could be less likely to occur in synthetic super-resolution datasets. Compared to U-Net, AEANets can perform better denoising, as shown in Figure~\ref{fig:vis_main}.

\subsection{The 3D FM Image Datasets}
We further evaluate our methods on Planaria and Flywing, two 3D microscopy transformation datasets from CARE~\cite{weigert2018content}. 
The Planaria dataset evaluates the model performance on the content-aware 3D image denoising task. It includes 17,005 pairs of noisy-clean 3D image patches of shape $64\times64\times16$ for training. The test data consists of 20 larger 3D images of size $1024\times1024\times95$. For each test image, the ground truth image and noisy images at three different noise levels captured under different lighting conditions (C1, C2, and C3) are provided. We evaluate our model on all the three noise levels.
The Flywing dataset evaluates the model performance on the content-aware 3D-to-2D image projection. Given a noisy 3D image, the projection task requires the transformation model to predict the surface of an organism in the 3D image and project it into a 2D images, excluding the noise along the depth dimension. The Flywing dataset consists of 16,891 pairs of noisy 3D and clean 2D patches for training and 26 test images. Similar to the Planaria dataset, each test image contains the ground truth version and the noisy version at three different noise levels.

Following the baseline configuration~\cite{wang2020global}, we apply our augmented attention operators to both the bottom block and up-sampling blocks for CARE datasets. For the 3D-to-2D projection task, we follow~\cite{weigert2018content, wang2020global} for the base model consisting of a 3D U-Net for surface projection followed by a 2D U-Net that further performs denoising on the projected image. Augmented attention blocks are included in both 3D and 2D U-Nets. As the 3D images for training are already large enough in their spatial size due to the additional depth dimension, we only apply the shared references and omit the batch-aware attention during training to avoid memory issue. \revision{In particular, the computational cost of an attention operator for a 3D input of spatial size $w\times h\times d$ can be $O(d^2)$ times the cost of a 2D input of spatial size $w\times h$.}

\begin{figure*}[th]
    \centering
    \includegraphics[width=0.98\textwidth]{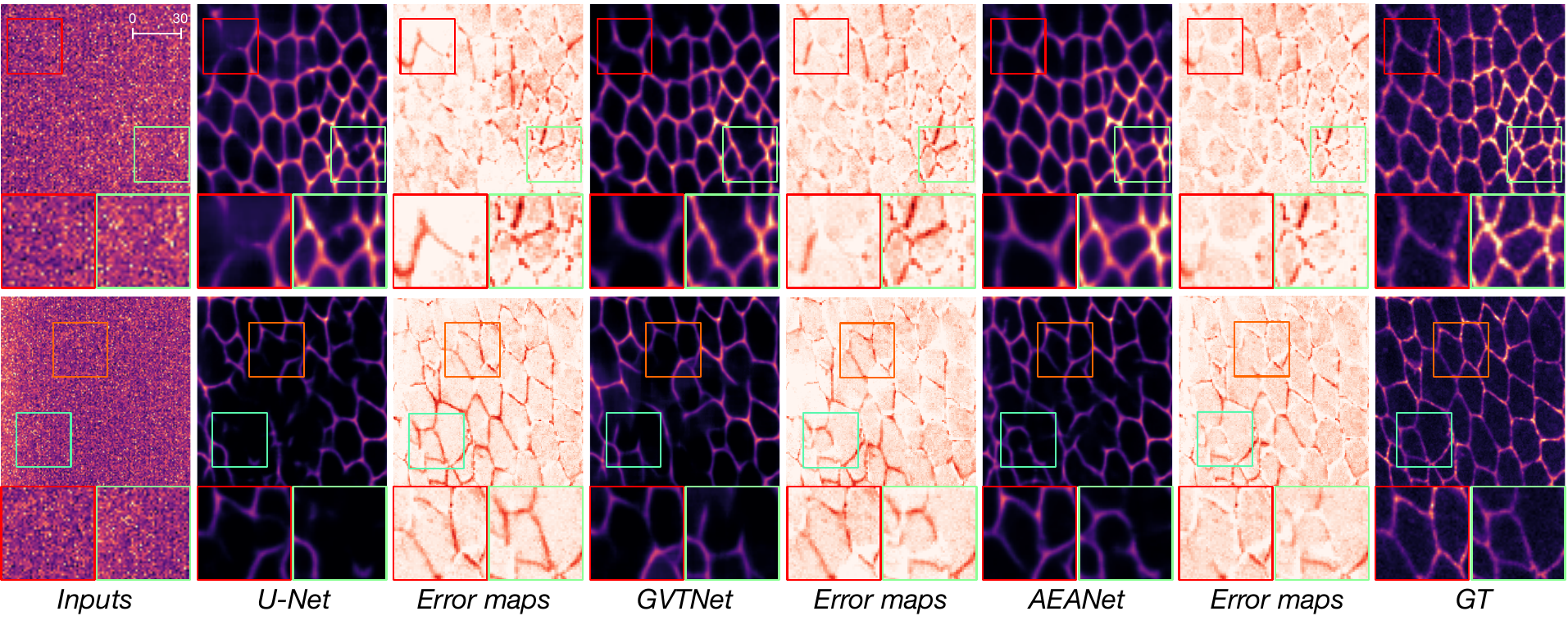}
    \caption{Visualization of the predicted 2D surface projections from the Flywing dataset. From left to right, the columns are the projection from input noisy volume, the predictions and error maps of U-Net, GVTNet, AEANet (ours), respectively, and the ground truth images. The images are predicted from noisy images with the worst lighting condition (C3). We zoom in some subareas for a better view.}
    \label{fig:vis_care}
    \vspace{-8pt}
\end{figure*}

The evaluation results in terms of PSNR and SSIM are shown in Table~\ref{tab:planaria} and Table~\ref{tab:flywing}. We include the evaluation metrics computed on the input images, and the predictions of the baseline methods, the U-Net and the GVTNet. We also include visualizations of the prediction results of three methods on the Flywing dataset in Figure~\ref{fig:vis_care}. The shown predictions are performed on noisy images with the worst lighting condition (C3). Both quantitative and visual results show that the proposed AEANet further consistently outperforms the current state-of-the-art methods on a wider range of microscopy image transformation tasks, indicating the effectiveness of the proposed augmented attention blocks.

\begin{table*}[]
\center
\caption{Evaluation results on the Planaria dataset for 3D image denoising. For all the three noise levels (C1, C2, and C3), model performance in terms of SSIM and PSNR are provided. The standard errors are computed among test samples following previous studies. The averaged scores over the three levels are also provided.}
\label{tab:planaria}
\begin{tabular}{lcccccc|cc}
\toprule
       & \multicolumn{1}{l}{C3 (SSIM)} & \multicolumn{1}{l}{C3 (PSNR)} & \multicolumn{1}{l}{C2 (SSIM)} & \multicolumn{1}{l}{C2 (PSNR)} & \multicolumn{1}{l}{C1 (SSIM)} & \multicolumn{1}{l|}{C1 (PSNR)} & \multicolumn{1}{l}{Avg (SSIM)} & \multicolumn{1}{l}{Avg (PSNR)} \\ \hline
Input  & 0.1561                        & 21.43                         & 0.1827                        & 21.73                         & 0.2260                        & 22.22                      & 0.1883                        & 21.79                          \\
Unet   & 0.6441$\pm$0.1207             & 28.13$\pm$1.37                & 0.7397$\pm$0.0885             & 30.15$\pm$1.66                & 0.7707$\pm$0.0889             & 31.57$\pm$1.71             & 0.7182                        & 29.95                          \\
GVTNet & 0.6972$\pm$0.1177             & 28.63$\pm$1.42                & 0.7745$\pm$0.0886             & 30.88$\pm$1.65                & 0.7929$\pm$0.0824             & 31.95$\pm$1.64             & 0.7549                        & 30.49                          \\
AEANet & \textbf{0.7073$\pm$0.0994}    & \textbf{28.64$\pm$1.39}       & \textbf{0.7764$\pm$0.0897}    & \textbf{30.95$\pm$1.84}     & \textbf{0.7933$\pm$0.0838}    & \textbf{32.08$\pm$1.70}             & \textbf{0.7590}               & \textbf{30.56}                 \\ \bottomrule
\end{tabular}
\end{table*}

\begin{table*}[]
\centering
\caption{Evaluation results on the Flywing dataset for 3D-to-2D image projection. For all the three noise levels (C1, C2, and C3), model performance in terms of SSIM and PSNR are provided. The standard errors are computed among test samples following previous studies. The averaged scores over the three levels are also provided.}
\label{tab:flywing}
\begin{tabular}{lcccccc|cc}
\toprule
       & \multicolumn{1}{l}{C3 (SSIM)} & \multicolumn{1}{l}{C3 (PSNR)} & \multicolumn{1}{l}{C2 (SSIM)} & \multicolumn{1}{l}{C2 (PSNR)} & \multicolumn{1}{l}{C1 (SSIM)} & \multicolumn{1}{l|}{C1 (PSNR)} & \multicolumn{1}{l}{Avg (SSIM)} & \multicolumn{1}{l}{Avg (PSNR)} \\ \hline
Input  & 0.0241                        & 16.62                         & 0.0795                        & 17.23                         & 0.1902                        & 18.38                      & 0.0979                         & 17.41                          \\
Unet   & 0.5592$\pm$0.0403             & 21.96$\pm$0.48                & 0.5971$\pm$0.0705             & 22.55$\pm$1.14                & 0.6067$\pm$0.0216             & 23.66$\pm$0.26            & 0.5877                         & 22.72                          \\
GVTNet & 0.5908$\pm$0.0465             & 22.36$\pm$0.43                & 0.6954$\pm$0.0248             & 24.28$\pm$0.38                & 0.7511$\pm$0.0257             & 25.81$\pm$0.33             & 0.6791                         & 24.15                          \\
AEANet & \textbf{0.6008$\pm$0.0452}      & \textbf{22.50$\pm$0.43}       & \textbf{0.7074$\pm$0.0305}      & \textbf{24.54$\pm$0.41}       & \textbf{0.7600$\pm$0.0195}      & \textbf{26.03$\pm$0.31}             & \textbf{0.6894}                & \textbf{24.36}                 \\ \bottomrule
\end{tabular}
\end{table*}

\subsection{3D Brain MRI Segmentation}
\revision{To demonstrate the effectiveness of AEANets on medical images and additional tasks, we further perform the evaluation on the 3D multimodality isointense infant brain MR image (MRI) dataset~\cite{wang2019benchmark}. The MRI segmentation task aims to segment given MR images by identifying different regions including cerebrospinal fluid (CSF), gray matter (GM), and white matter (WM) regions. The MRI segmentation is also considered as an image transformation task as it performs pixel-wise classification and hence requires the model to be spatially permutation equivariant.}

\revision{We follow \cite{wang2020non} for the network, training configuration, and evaluation setting only except for the AEA block. In particular, we perform the leave-one-subject-out cross-validation on the ten public MRI subjects and compute the Dice ratio as the evaluation metric. The results shown in Table~\ref{tab:mri} indicate that AEANets achieve consistently better performance compared to the close baseline Non-local U-Net.}

\begin{table*}[]
\centering
\caption{\revision{Evaluation results in terms of Dice Ratios on the 3D brain MRI segmentation task. The 10-fold cross-validation is adopted to compute the scores.}}
\label{tab:mri}
\begin{tabular}{lccc|c}
\toprule
Model           & CSF Dice Ratio & GM Dice Ratio & WM Dice Ratio & Avg. Dice Ratio\\\hline
CC-3D-FCN~\cite{nie20183}       & 0.9250±0.0118 & 0.9084±0.0056 & 0.8926±0.0119 & 0.9087±0.0066 \\
Non-local U-Net~\cite{wang2020non} & 0.9530±0.0074 & 0.9245±0.0049 & 0.9102±0.0101 & 0.9292±0.0050 \\
AEANet         & \textbf{0.9556±0.0062} & \textbf{0.9279±0.0052} & \textbf{0.9136±0.0117} & \textbf{0.9324±0.0052} \\\bottomrule
\end{tabular}
\end{table*}

\subsection{Verification of Equivariance Properties}
\begin{figure}[h]
    \centering
    \includegraphics[width=0.48\textwidth]{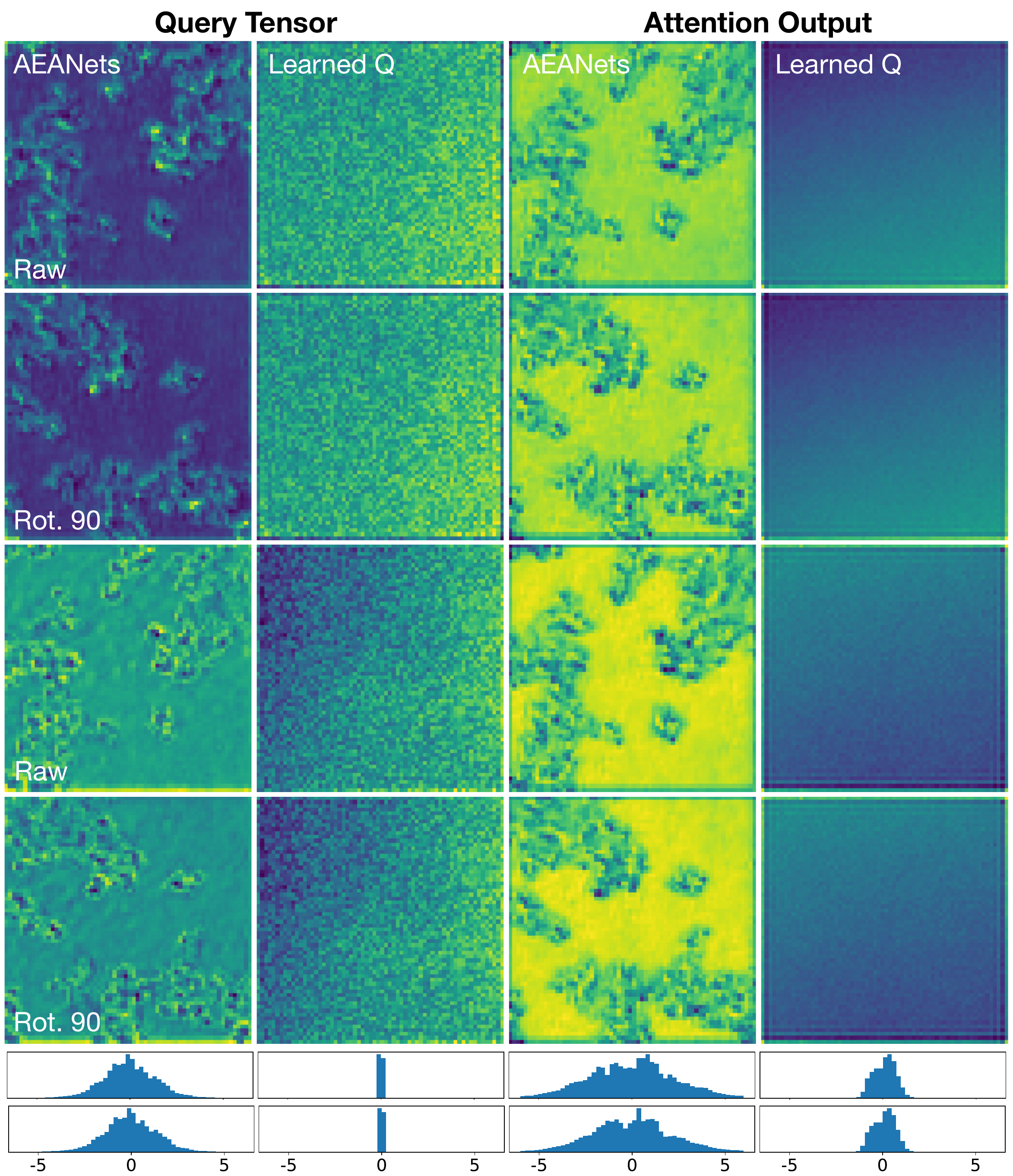}
    \caption{\revision{Visualization of query tensors and attention output tensors of AEANets and Attention with learned queries. Each row corresponds to one channel (out of many) of the tensors. At the bottom are the distributions of values in the tensors with raw and rotated input, respectively.}}
    \label{fig:att_eq}
    \vspace{-8pt}
\end{figure}

\scndrev{To better support our theroy and claims, we empirically verify the spatial permutation equivariance property of AEANets and attention with learned query. In particular, we visualize the first two channels of the query tensor and attention outputs when the raw patch and a rotated patch are input to the model.} \revision{The visualizations are shown in Figure~\ref{fig:att_eq}. For the attention block with learned query, the values in both query and attention output tensor reduce to nearly constant among spatial locations (according to the histograms) and do not rotate accordingly with the input patch. This is due to the permutation invariance nature of attention block with learned query, who becomes permutation equivariance if and only if the learned query reduces to constant values at all locations. However, in this case, the learned query is unable to capture common features among the dataset and hence becomes meaningless.} \scndrev{The results also indicate the issue related to the invariance property cannot be addressed by data augmentations.} In contrast, the AEA block is able to remain spatially permutation equivariance according to the visualization while capturing common features.

\revision{On the model level, we quantitatively evaluate the equivariant property by computing the difference, in terms of mean absolute error, between the outputs of raw and permuted input patches. We evaluate models with the self-attention block, the attention with learned query, and the proposed AEA block under rotations and transpose permutations. MAE scores in Table~\ref{tab:equv} demonstrate that the proposed AEA block can achieve even better permutation equivariance compared the self-attention block. Note that although the attention with learned query can also achieve a similar level of equivariance, it is unable to learn meaningful query tensors or outputs as the query tensor reduces to constant values over spatial locations.}

\begin{table}[]
\centering
\caption{Quantitative evaluation of the equivariance to spatial permutations . Shown are mean absolute errors (MAEs) between outputs for raw and permuted input patches.}
\label{tab:equv}
\begin{tabular}{lcccc}
\toprule
Methods & Rot. 90 & Rot. 180 & Rot. 270 & Traspose \\ \hline
Self-attention    & 0.01529 & 0.02035  & 0.01518  & 0.00810  \\
Learned query & \textbf{0.01322} & 0.01757  & 0.01369  & 0.00712  \\
AEA block   & 0.01351 & \textbf{0.01689}  & \textbf{0.01358}  & \textbf{0.00506}  \\ \bottomrule
\end{tabular}
\end{table}

\subsection{Ablation Studies}

We conduct an ablation study to analyze (1) how the shared references and the Batch-aware Attention mechanism help improve the performance of attention-based models and (2) how AEANets benefit from a larger input image size. For a fair comparison, all the models in this subsection are trained with pooled-training.

We first evaluate the performance of AEANets with the following options: Batch-aware Attention excluded, shared references excluded, and shared references with different sizes (16, 32 and 64). The results in terms of $\Delta$PSNR and $\Delta$SSIM are shown in Table~\ref{tab:ablation}. Compared to the original attention-based model, GVTNets, applying Batch-aware Attention and shared references renders a performance gain of 0.09 dB and 0.17 dB, respectively. When increasing the size of the shared references, the performance of AEANets also increases.

\begin{table}[th]
\centering
\caption{Performance of AEANets when Batch-aware Attention is excluded or the size of shared references are decreased. Three baseline methods are also given for comparison.}\label{tab:ablation}
\begin{tabular}{lcc}
\toprule
Methods                     & $\Delta$PSNR (dB) & $\Delta$SSIM \\ \hline
RCAN                        & 1.59         & 0.051        \\
U-Net                       & 1.46         & 0.074        \\
GVTNets              & 1.87         & 0.086        \\ \cline{1-3}
Shared Reference (SR) only  & 2.04         & 0.085        \\
Batch-Aware (BA) only    & 1.96         & 0.084        \\
BA + SR (16)                 & 1.98         & 0.085        \\
BA + SR (32)                 & 2.02         & 0.085        \\
BA + SR (64)                 & \textbf{2.10}         & \textbf{0.087}        \\ \bottomrule
\end{tabular}
% \vspace{-8pt}
\end{table}

Regarding the size of input images, we evaluate AEANets on the same testing images but with different input sizes. In particular, we crop each image into patches of a given size, input the patches to the network and then stitch the predicted patches together as the prediction of the entire image. We evaluate the improvement in PSNR for each size and show the results in Supplementary Figure~\ref{fig:over_patch}. Among the evaluated alternative, both GVTNet and AEANets benefit from a larger patch size since both are attention-based models. It is interesting to see that the performance of GVTNet with full-sized input can be achieved by AEANets with much smaller input patch sizes. Specifically, the AEANet with shared references of size 64 and the input size 256 reaches the similar performance of the GVTNet with full-sized input.

\section{Conclusion}
High-quality microscopy images in terms of resolution or noise level are usually desired for better Biomedical and Nanomaterial researches. Computational methods that perform super-resolution and denoising on microscopy images make it possible to obtain high-quality microscopy images more efficiently with lower cost. In this work, we consider the microscopy image-to-image transformation and focus on challenges in the case where both high-quality and low-quality images in the training dataset are physically captured. To address the challenges, we have introduced the Augmented Equivariant Attention Networks (AEANets), which is able to utilize shared features among images and inter-image dependencies, and preserve the spatially permutation equivariant property for image-to-image transformation. We have theoretically analysed the property of the proposed attention operator augmented by shared references and the property of existing attention operators as comparisons. And we have conducted experiments to show the effectiveness of AEANets.

% \section*{Acknowledgments}
% The authors would like to acknowledge the generous support from their sponsors.
% \vspace*{-12pt}

% \begin{thebibliography}{}
\bibliographystyle{IEEEtran}
\bibliography{reference,ji}

\clearpage
\appendices
\setcounter{page}{1}

\section{Discussion of the invariance and equivariance properties}\label{supp:disc}

We consider the properties under three common spatial permutation cases, \emph{i.e.}, rotation, flipping and shifting.

In a natural image classification task, when the input image is rotated, flipped or shifted, we expect the classification result to remain the same as long as the object to be classified is still in the image, as shown in Figure~\ref{fig:inv_eqv} (left). In this case, the model can benefit from its invariance property and an operator that is spatially permutation invariant can help the model realize such property and hence improve its generalization capability.

On the contrary, when performing the rotation, flipping or shifting on the input of an image-to-image transformation model, we expect the output image of the model to be rotated, flipped or shifted correspondingly, as shown in Figure~\ref{fig:inv_eqv} (right). Hence the equivariance to spatial permutation is desired by the model. In this case, if a spatially permutation invariant operator is included, the equivariance will be violated, since the operator outputs a constant tensor while the input image is rotated, flipped or shifted. Hence, operators with such an invariant property can be inappropriate in image-to-image transformation models and may lead to a performance reduction. It is desirable to use a spatially permutation equivariant operator to preserve the equivariance of the model.

\section{Proof of Theorem 1}\label{supp:proof1}
\begin{proof}
When applying a spatial permutation $\mathcal{T}_\pi$ to the input $\bm X$ of a self-attention operator $A_s$, we have
\begin{equation}
\begin{aligned}
    A_s(\mathcal{T}_\pi(\bm X)) &= \left(\frac{1}{s}\mathcal{T}_\pi(\bm Q)\cdot(\mathcal{T}_\pi (\bm K))^T\right)\cdot\mathcal{T}_\pi (\bm V) \\
    &= \frac{1}{s}P_\pi \bm Q\bm K^T(P_\pi^TP_\pi)\bm V \\
    &= P_\pi \left(\frac{1}{s}\bm Q\bm K^T\right)\bm V \\
    &= \mathcal{T}_\pi(A_s(\bm X)).
\end{aligned}
\end{equation}
Note that $P_\pi^T P_\pi=I$ since $P_\pi$ is an orthogonal matrix. Since convolutions with a kernel size of $1$ are permutation equivariant, the projected $\bm Q=q(\bm X), \bm K=k(\bm X), \bm V=v(\bm X)$ are spatially permutation equivariant with respect to the input $\bm X$. By showing $A_s(\mathcal{T}_\pi(\bm X))=\mathcal{T}_\pi(A_s(\bm X))$ we have shown that $A_s$ is spatial permutation equivariant according to Definition 2.

In comparison, when applying $\mathcal{T}_\pi$ to the input of an attention operator $A_Q$ with a learned query $\bm Q$, which is independent of the input $\bm X$, we have
\begin{equation}
\begin{split}
    A_{\bm Q}(\mathcal{T}_\pi(\bm X)) &= \left(\frac{1}{s}\bm Q\cdot(\mathcal{T}_\pi (\bm K))^T\right)\cdot\mathcal{T}_\pi (\bm V) \\
    &= \frac{1}{s}\bm Q\bm K^T(P_\pi^TP_\pi)\bm V \\
    &= \left(\frac{1}{s}\bm Q\bm K^T\right)\bm V \\
    &= A_{\bm Q}(\bm X).
\end{split}
\end{equation}
Since $A_{\bm Q}(\mathcal{T}_\pi(\bm X))=A_{\bm Q}(\bm X)$, we have shown that $A_{\bm Q}$ is spatial permutation invariant according to Definition 2.%\hfill\qedsymbol
\end{proof}

\section{Proof of Theorem 2}\label{supp:proof2}
\begin{proof}
We let
$
    \tilde P_{\pi} =
    \left(\begin{smallmatrix}
    \bm I_{r} & \bm 0\\
    \bm 0 & P_\pi
    \end{smallmatrix}\right),
$
where $P_\pi$ is the permutation matrix applied to $\bm X$. Then we have
\begin{equation}
\begin{split}
    A_{\bm R}(\mathcal{T}_\pi(\bm X)) &= \left(\frac{1}{r+wh}\mathcal{T}_\pi(\bm Q)\cdot\left(\tilde P_\pi \tilde{\bm K}\right)^T\right)\cdot\tilde P_\pi\tilde{\bm V} \\
    &= \frac{1}{r+wh}P_\pi \bm Q\tilde{\bm K}^T(\tilde P_\pi^T\tilde P_\pi)\tilde{\bm V} \\
    &= P_\pi \left(\frac{1}{r+wh}\bm Q\tilde{\bm K}^T\right)\tilde{\bm V} \\
    &= \mathcal{T}_\pi\left(A_{\bm R}(\bm X)\right).
\end{split}
\end{equation}
This shows that the proposed attention operator augmented with shared references is spatially permutation equivariant.%\hfill\qedsymbol
\end{proof}

% \newpage
\section{Network Architecture}\label{supp:network}
\begin{figure}[th]
    \centering
    \includegraphics[width=0.45\textwidth]{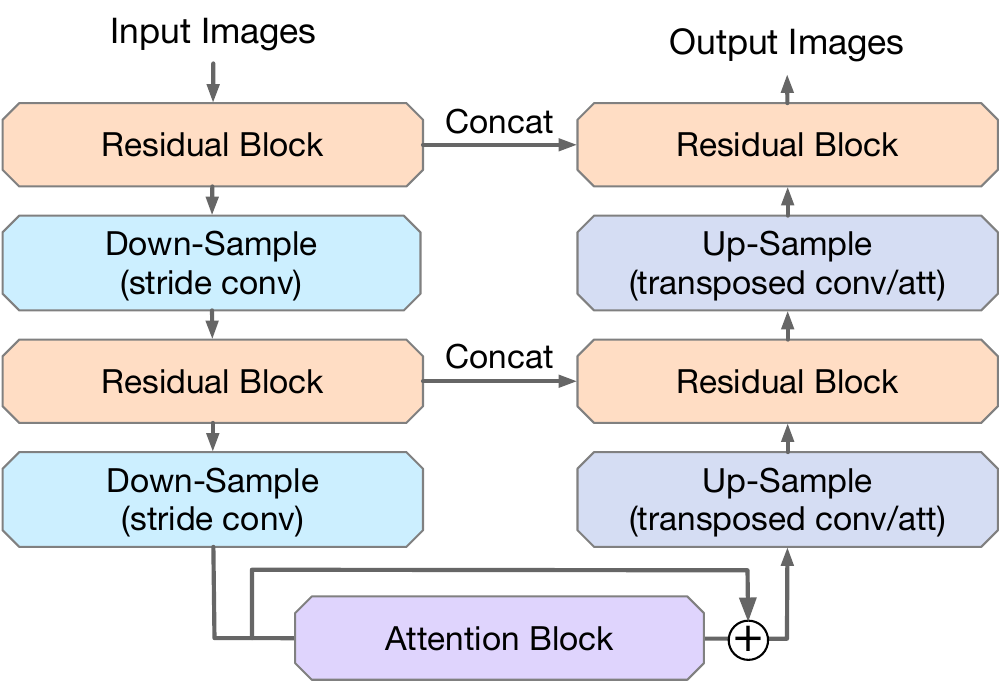}
    \caption{Network Architecture. We chip our proposed attention block into a U-Net architecture with depth of 3. The skip-connections in the U-Net use concatenation. The down-sampling applies the stride convolution and the up-sampling applies either the transposed convolution or the augmented attention block.}
    \label{fig:network}
\end{figure}

The overall architecture of the proposed network is shown in Figure~\ref{fig:network}.

\section{Implementation Details and Configurations}\label{supp:config}
\revision{For all three learning tasks on the four datasets, we implement our methods using TensorFlow 1.14 and perform training and testing on a single NVIDIA GeForce RTX 2080 Ti GPU. Below are specific configurations for individual datasets.}

\revision{For image super-resolution on the Paired 2D EM dataset, we adopt a U-Net with depth of 3 (including 2 down-sampling and 2 up-sampling) as the base structure and replace the bottom block by the proposed AEA block with batch-aware training. The training patches are randomly sampled from training subimages and cropped into a size of $256\times256$. We adopt a mini-batch size of 8, where 4 samples go through the attention with shared references and the other 4 samples go through the batch-augmented attention. The two attention operators have their parameters, \textit{i.e.}, projection weights for query, key, and value, shared by setting \textit{reuse=True} at implementation. The model adopts the mean absolute loss as the learning objective and is optimized with Adam optimizer under a base learning rate of $0.0004$ with an exponential learning rate decay by $0.5$ for every $10,000$ steps. The model is trained for a total of $120,000$ training steps.}

\revision{For 3D FM image restoration and MR image segmentation, we follow previous works, \cite{wang2020global} and \cite{wang2020non}, respectively, for most model configurations and training settings including the depth of network, type of convolutions, training patch sizes, etc. Models for Planaria restoration and Flywing projection are trained for $80$ epochs with batch sizes of 16. As the previous work~\cite{wang2020global} adopts self-attention blocks at both bottom block and up-sampling blocks, we replace all self-attention blocks by the proposed AEA block for a fair comparison and to better show the effectiveness of the AEA block. For the MR image segmentation, the only difference with the previous work Non-local U-Nets~\cite{wang2020non} is to replace the attention block by our AEA block. Other network configurations and training settings are kept the same. Models for all ten folds are trained for $300,000$ steps.}

\revision{According to a recent study~\cite{venkataramanan2021hitchhiker}, the scores computed by different implementations of the SSIM metric may vary. For fair comparisons, we closely follow individual baseline works for the SSIM implementations of each experiment. In particular, for the CARE 3D FM experiments, we use the original evaluation code~\cite{weigert2018content} based on the scikit-image implementation. For the 2D EM experiments, we adopt a matched implementation of Wang et al.~\cite{wang2004image} (with gaussian weights, sigma=1.5, and covariance sampling disabled) following the previous study~\cite{qian2020effective}.}

\section{Effects of Shared References}

\begin{figure*}[h]
    \centering
    \includegraphics[width=0.98\textwidth]{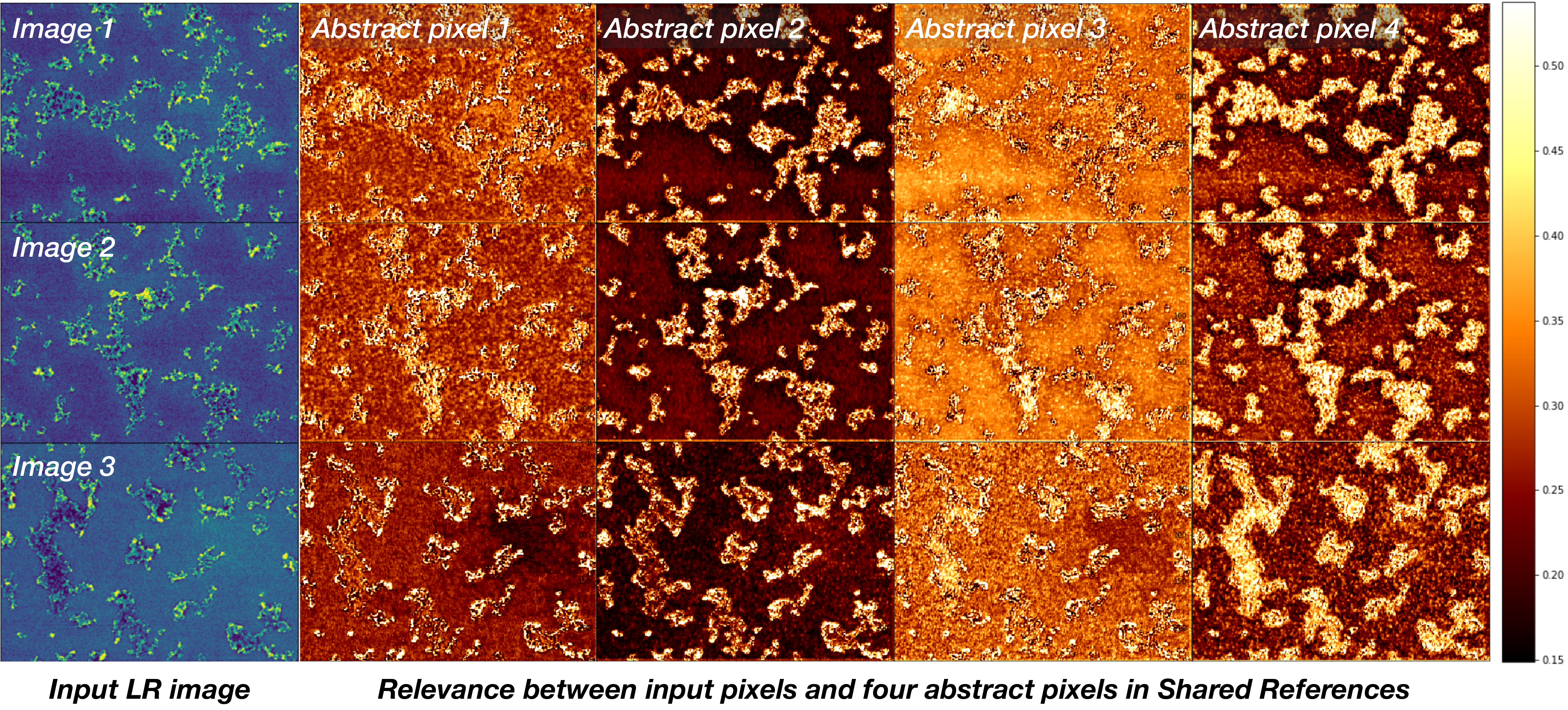}
    \caption{The visualization of the relevance between pixels in the input image and four randomly selected abstract pixels from shared references on the Paired EM Image dataset. From top to bottom, rows are the visualizations of different input images. From left to right, the first column shows the input images and the rest four columns are the visualizations for the four selected abstract pixels. A higher value in the heatmap indicates stronger relevance.}
    \label{fig:att_visual}
\end{figure*}

This subsection provides a discussion on the effects o shared references.  Recall that in the learned shared references $\bm R=[\bm f_1,\cdots, \bm f_r]^T\in\mathbb{R}^{r\times c}$ of size $r$, we call each row vector $\bm f_i\in\mathbb{R}^{1\times c}$ the feature vector of an abstract pixel distilled from the training images. To illustrate the effects of the shared references, we select three input images and randomly select the feature vectors of four abstract pixels.  We visualize how much the Query matrix $\bm Q$ of the three images is correlated to the four abstract pixels. Provided the Query matrix $\bm Q_i\in\mathbb{R}^{wh\times c_1}$ of an input image and the feature vector $f_j$ of an abstract pixel, we visualize
$$\bm Q_i\cdot k^T(f_j)\in\mathbb{R}^{wh\times1},\;i\in\{1,2,3\},\;j\in\{1,2,3,4\},$$
where $k(\cdot)$ projects the feature vector into the Key space and the dot product indicates the relevance between each pixel in the input image and the learned abstract pixel. We fold $\bm Q_i\cdot k^T(f_j)$ back to the original 2D spatial shape $w\times h$ and visualize it in Figure~\ref{fig:att_visual} for each $(i,j)$ in the form of a heatmap. The visualization shows which pixels (or segments) in the input image are tightly related to a given abstract pixel in the shared references.

The four columns on the right show different patterns, suggesting that the abstract pixels in the shared references contain different types of features and are related to different segments of the input images. The abstract pixels in the same column show similar patterns, indicating that features captured by the abstract pixels are shared across images. These two observations tell us that the effect of the shared references matches our expectation.

\section{Results of ablation on training patch sizes}

\begin{figure}[H]
    \centering
    \includegraphics[width=0.45\textwidth]{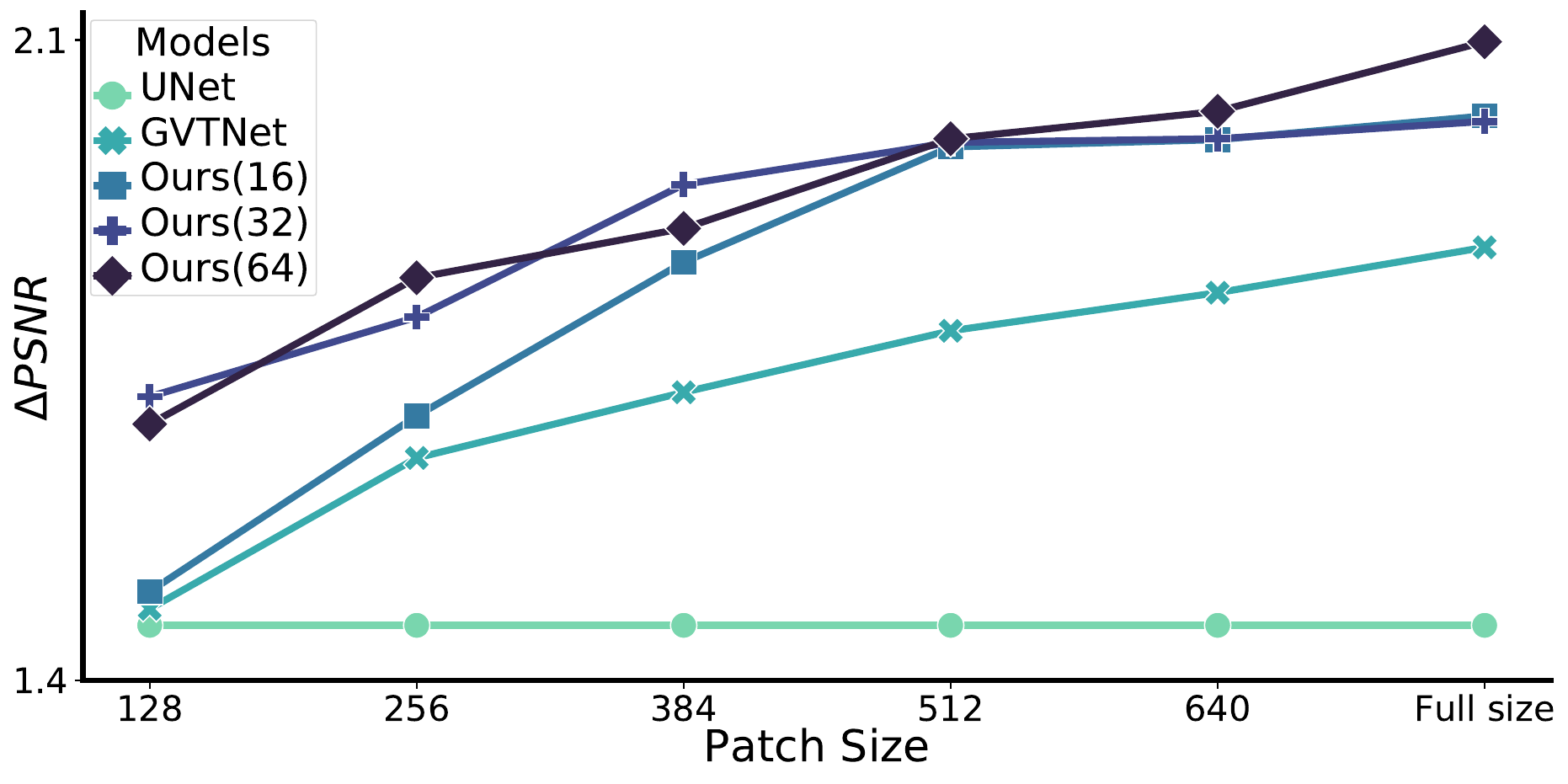}
    \caption{The output image quality $\Delta$PSNR over different input patch sizes. Results are computed on testing images. Numbers inside the brackets in models, \emph{i.e.}, 16, 32 and 64, refer to the sizes of shared references. For all attention-based methods (GVTNet and ours), the output image quality increases when larger input patches are given.}
    \label{fig:over_patch}
\end{figure}

Change of performance over different training patch sizes is shown in Figure~\ref{fig:over_patch}.

\section{Discussions on Limitations and Future Directions}

\revision{\paragraph{Temporal Super-resolution}
When capturing a series of microscopy images as a temporal sequence, one has to trade-off between the spatial resolution or quality of each frame and the temporal resolution (fps). Such limitation can be also addressed by extending the advanced image transformation approaches. Besides capturing the sequence in high temporal resolution and computationally obtain high-quality frames, one can also perform temporal super-resolution to directly improve the temporal resolution. The latter case is also an image-to-image transformation problem when considering the temporal dimension as an additional spatial dimension. In both cases, the transformation can benefit from additional information along the temporal dimension, \textit{e.g.}, by aggregating temporal information with attention operators. However, the attention operators requires careful design to enable efficient computation as the temporal dimension brings significantly higher computational cost.
}

\revision{\paragraph{Transformation Equivariance with Anisotropic Images}
For 3D microscopy, it is common that anisotropic images are captured, where the resolution along the Z-axis (depth) is inconsistent with the other axes. In this case, it is more challenging to achieve spatial transformation equivariance.} \scndrev{We do not include experiments related to the anisotropic problem as it is not part of our claims or conclusions.} \revision{While our work does not aim at addressing the anisotropic issue of 3D images, the attention-based operators (including the AEA block) can be a promising solution to the anisotropic issue. As the attention-based operators perform non-local aggregation among voxels, a permutation on the input such as transpose that exchanges the resolution of two axes (or spatial distortions) does not change the final output value at corresponding spatial locations. However, the bottleneck of addressing the anisotropic issue lies in the convolutional operators, who rely on preset resolutions along different axes, and performing transpose to the input of such operators will change its outputs. Potential solutions on this issue include cooperating with gating mechanisms to identify inconsistency in resolutions or to adopt fully attention-based networks.}

\end{document}